\documentclass[runningheads]{llncs}
\usepackage[T1]{fontenc}
\usepackage{graphicx}
\usepackage{amsmath}
\usepackage{url}
\usepackage{subcaption}
\usepackage{graphicx}
\usepackage{microtype}

\begin{document}
\title{Beyond Collision Avoidance: Multi-Robot Yielding and Spatial Affordance in Emergency Evacuations}
\titlerunning{Beyond Collision Avoidance: Multi-Robot Emergency Evacuation Study}
%
\author{Ning Zhou\inst{1}\orcidID{0009-0005-8352-1841} \and
Edmund R. Hunt\inst{1}\orcidID{0000-0002-9647-124X} \and
Nikolai W.F. Bode \inst{1}\orcidID{0000-0003-0958-5191}}
\authorrunning{N. Zhou et al.}
%
\institute{School of Engineering Mathematics and Technology, University of Bristol,
Bristol, UK\\
\email{\{ning.zhou,edmund.hunt,nikolai.bode\}@bristol.ac.uk}}
\maketitle              
\begin{abstract}
As mobile service robots increasingly coexist with pedestrians, ensuring \textit{passively safe} behaviour during confined emergency evacuations is critical. Existing multi-robot yielding strategies often focus solely on collision avoidance and macroscopic flow optimisation, overlooking environmental affordances and human spatial expectations. To bridge the gap between macroscopic theory and micro-level perception, we conducted a game-based virtual evacuation experiment (N=56). We investigated individual psychological responses to four multi-robot yielding strategies (\texttt{Hide}, \texttt{LineEscape}, \texttt{Freeze}, \texttt{ShortestPath}) across confined corridors with and without refuge niches. Our results establish a robust preference hierarchy (\texttt{Hide} $>$ \texttt{LineEscape} $>$ \texttt{Freeze} $>$ \texttt{ShortestPath}), demonstrating that proactive space-yielding significantly outperforms freezing and efficiency-first approaches. Crucially, we found that environmental affordances heavily shape cognitive expectations. Actively utilising available niches amplifies the psychological comfort of proactive yielding (\texttt{Hide}). Conversely, failing to use an obvious niche (e.g., executing \texttt{LineEscape}) may trigger Expectation Violation. This is reflected in a drastically increased perceived cognitive delay, despite objectively unimpeded trajectories. Furthermore, prior robot interaction experience helps users decode complex social intents. Ultimately, this research demonstrates that safe human-robot interaction during emergencies must evolve from pure trajectory optimisation to semantically aware navigation. Future work will extend this framework to investigate complex interactions between robot swarms and pedestrian crowds.

\end{abstract}

\section{Introduction}
\label{sec:intro}

As mobile robots increasingly populate public spaces, their presence during emergency evacuations poses a significant safety concern. While often envisioned as active evacuation assistants~\cite{bahamid2020}, legal liabilities and complex trust dynamics largely preclude robots from assuming authoritative \textit{guiding} roles. In high-stakes scenarios, autonomous robots that fail to meet human expectations cause trust violations that take time to repair, even when explanations are provided~\cite{webb2025}. Consequently, the industry default fallback remains a \textit{collective freeze} when encountering crowds~\cite{trautman2010}. However, while freezing is safe for an isolated robot, a stationary swarm in a confined corridor physically impedes evacuees. This highlights a critical, under-explored imperative: designing multi-robot systems with \textit{passively safe} behaviours that minimise spatial and psychological interference without disrupting human evacuation flow.

Even within the pursuit of passive safety (e.g., robot yielding cues~\cite{hetherington2021}), existing research exhibits a significant gap. While recent advancements in social navigation have successfully integrated \textit{spatial semantics}---the social rules of a space, such as knowing which areas are for walking versus waiting~\cite{moller2021}---these approaches are rarely applied to time-critical emergency evacuations. Under evacuation pressures, the role of \textit{environmental affordances} has been largely overlooked. An affordance is a physical feature that naturally suggests a specific action: for instance, a side-niche in a corridor invites an agent to step out of the main flow~\cite{gibson1977}. In a real-world corridor, architectural features like refuge niches, utility rooms, or opened doors provide these clear opportunities for yielding. When such spaces exist, human expectations undergo a shift: the physical niche becomes a social obligation for the robot. If a robot merely optimises its trajectory but fails to proactively utilise these obvious spots, it can trigger an \textit{Expectation Violation}~\cite{asavanant2021,burgoon1993}. This mismatch between human spatial intuition and robot behaviour can increase users' perceived cognitive load and subjective sense of delay, even if their physical path remains objectively unimpeded.

While the ultimate challenge of emergency navigation involves complex crowd dynamics, macroscopic models often treat humans as interacting particles, stripping away individual psychological comfort and spatial expectations~\cite{chatagnon2025}. Before introducing the confounding variables associated with multiple interacting pedestrians, establishing a foundational baseline of micro-level human-robot spatial negotiation is essential. By isolating interactions to a single individual, we conducted a game-based simulation evaluating human responses across two architectural layouts (\texttt{No Affordance} vs. \texttt{Affordance}) and four navigation behaviours: \texttt{Hide}, \texttt{ShortestPath}, \texttt{Freeze}, and \texttt{LineEscape}. We present three key contributions: (1) a comparative evaluation of proactive versus efficiency-first yielding strategies in confined emergencies; (2) an analysis of how affordances shape spatial expectations and the psychological friction caused by failing to utilize niches; and (3) an investigation into how prior human-robot interaction experience helps users decode complex robot intentions.

\section{Related Work}
\label{sec:related_work}

\subsection{Robots in Emergency Evacuations: From Active Guiding to Passive Safety}

The deployment of mobile robots in emergency scenarios has traditionally focused on active intervention. Prior research has frequently envisioned robots as active guides~\cite{Nayyar2019}, dynamic signage~\cite{hetherington2021}, or regulating pedestrian flow~\cite{Zheng2022} that actively guide pedestrians toward exits~\cite{bahamid2020}. While active guidance strategies can be effective for disoriented individuals, recent studies highlight the severe legal liabilities, ethical concerns, and technical risks associated with granting robots authoritative roles during life-threatening crises. For instance, a well-mentioned technical risk is the issue of overtrust, where humans might blindly follow a malfunctioning robot into danger~\cite{robinette2016}. Consequently, this suggests a shift towards passively safe behaviours. In this context, robots are not expected to be actively guiding humans, but rather to behave in a socially acceptable manner that strictly minimises spatial interference and avoids impeding the human evacuation flow.

\subsection{Multi-Robot Yielding and Crowd Dynamics: The Physical Dimension}

Enabling passive safety in confined spaces requires effective yielding strategies. Here, we briefly introduce the four strategies examined (see also Section~\ref{sec:strategies}). For single-robot scenarios, spatial yielding cues are proven to increase robot legibility and human comfort~\cite{hetherington2021}. However, extrapolating these single-robot conventions to multi-robot swarms introduces critical challenges. The industry default is the \textit{Collective Freeze} (\texttt{Freeze}) strategy~\cite{trautman2010}, where robots stop upon detecting a crowd. While safe for an individual robot, a swarm freezing in a narrow corridor creates obstacles for passing pedestrians. A natural spatial optimisation is for robots to actively yield by pulling over to the corridor boundaries, theoretically justifying our \texttt{Hide} strategy. Yet, without architectural recesses, these yielded robots remain static obstacles in the corridor. In macroscopic crowd dynamics, such static bottlenecks can anchor force chains, structurally inducing a \textit{jamming effect} analogous to granular flow dynamics~\cite{seguin2022}.

To mitigate the congestion caused by static obstacles, macroscopic crowd dynamics research suggests several dynamic interventions to optimise flow. One approach suggests that obstacles moving along the boundaries with the pedestrian flow can mitigate jamming effects and maintain evacuation momentum~\cite{xu2022}, which theoretically justifies our \texttt{LineEscape} strategy. Alternatively, studies on pedestrian dynamics demonstrate that static or moving dividers positioned in the centre of a corridor can induce spontaneous line formation, adaptively shaping pedestrian traffic and effectively accelerating movement---a phenomenon particularly beneficial in bidirectional flows or multi-exit scenarios~\cite{Koyama2020}. This principle, combined with the objective of rapid self-evacuation to permanently free up spatial capacity for humans, provides the theoretical justification for our \texttt{ShortestPath} strategy, where robots navigate directly down the centre of the corridor. Setting aside psychological factors, these flow-compliant strategies should theoretically represent the optimum for macroscopic evacuation efficiency during medium- or high-density pedestrian flows.

\subsection{Environmental Affordances and Expectation Violation: The Psychological Dimension}

Despite the advantages of flow-compliant strategies, human-robot social navigation cannot be reduced to purely physical particle dynamics. In everyday scenarios, modern social navigation frameworks have already begun to move beyond trajectory optimisation by incorporating spatial semantics---enabling robots to adapt their behaviours based on room types, static furniture, or invisible social interaction spaces~\cite{moller2021,zhang2025}. However, a notable gap remains in emergency evacuation research, where macroscopic flow optimization models largely overlook these socio-physical semantics.

Confined corridors inherently force robots into a human's intimate or personal space, which, according to established theories of human-robot proxemics, elevates psychological discomfort and anxiety~\cite{gomez2013,hall1966}. Yet, corridors are rich in \textit{environmental affordances}---physical features of an environment that naturally invite or suggest specific actions~\cite{gibson1977}---such as physical refuge niches, vacant rooms, or widened intersections. When a robot leverages these affordances to physically remove itself from the user's immediate path, it effectively transitions away from the user's personal space to their social space~\cite{hall1966}. This decreased physical proximity naturally yields higher psychological comfort, explaining why the utilisation of physical niches fundamentally improves human perception.

Beyond physical proximity, the presence of spatial affordances is anticipated to fundamentally improve human cognitive expectations. Expectancy Violations Theory (EVT)~\cite{burgoon1993,burgoon1976} posits that humans form specific social expectations for robotic agents; when a robot violates these expectations without a clear justification, it leads to severe negative communication outcomes and decreased trust~\cite{asavanant2021}. When a clear physical yielding space exists, the human expects a socially competent robot to proactively utilise it. Failing to use this obvious affordance is likely to constitute negative expectancy violation, which may drastically degrades the user's subjective evaluation. This psychological friction explains why optimal flow algorithms may fail in real-world human evaluations, thereby necessitating the introduction of proactive, affordance-aware yielding strategies~\cite{bera2017,gao2022}.

\subsection{Research Questions}
To isolate the psychological impact of the robots' algorithms from the complex confounding variables of human-human crowd dynamics, this study deliberately focuses on micro-level, one-on-one spatial negotiation as a foundational baseline. Based on the conflicting priorities between physical dynamics and psychological spatial expectations, this study aims to answer the following research questions (RQs):
\begin{itemize}
    \item \textbf{RQ1:} How do different multi-robot yielding strategies dynamically affect human subjective perception and objective evacuation behaviour during one-to-many interaction in a confined corridor?
    \item \textbf{RQ2:} How do environmental affordances (i.e., the presence of refuge niches) moderate perceptions, and does the failure of a robot to utilise these affordances trigger behavioural and psychological indicators of expectation violation in single-person interactions?
\end{itemize}

\section{Methodology}
\label{sec:methodology}

\subsection{Virtual Evacuation Gameplay and Robot Strategies}
\label{sec:strategies}

\begin{figure}[h]
\centering
\includegraphics[width=1\textwidth]{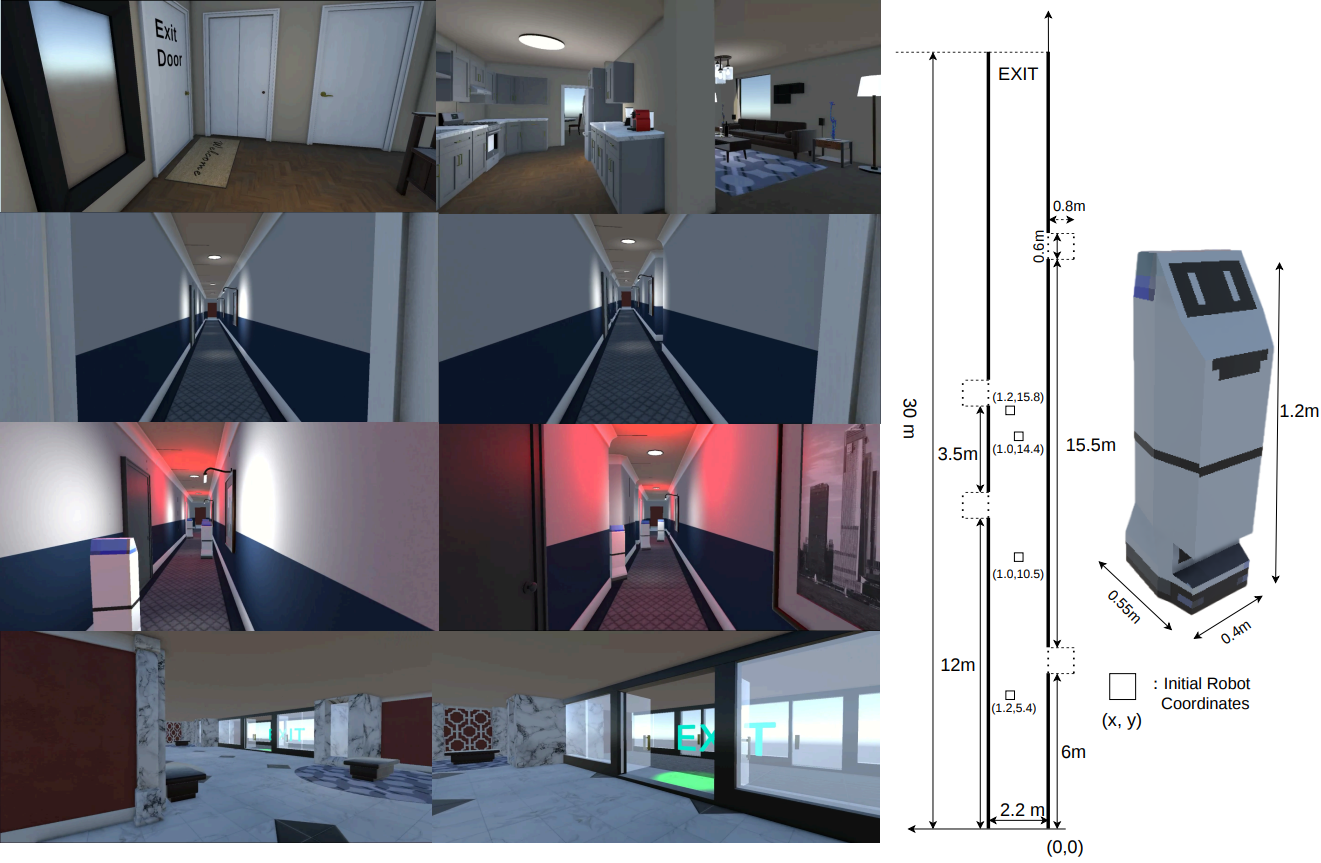}
\caption{First-person perspective of the virtual evacuation simulation pipeline (left) alongside the top-down architectural schematic and precise robot dimensions (right). Row 1: The participant's initial starting locations inside the hotel rooms. Row 2: Baseline training sessions in the \texttt{No Affordance} (\texttt{A0}, left) and \texttt{Affordance} (\texttt{A1}, right) corridors without robots. Row 3: Formal evacuation trials triggered by the red-flashing fire alarm, depicting the \texttt{LineEscape} strategy in \texttt{A0} (left) and the \texttt{Hide} strategy in \texttt{A1} (right). Row 4: The hallway and final exit doors.}
\label{fig:pipeline}
\end{figure}

To investigate the dynamics of human-robot spatial negotiation during an emergency, a first-person perspective evacuation simulation was developed using the Unity3D\footnote{\url{https://github.com/BlankNing/ICSR26HRI}} game engine. The simulation placed participants in a virtual hotel environment where they were required to evacuate from their room, navigate through a confined $2.2\,\text{m} \times 30\,\text{m}$ corridor as detailed in Fig.~\ref{fig:pipeline} right, and reach the safety exit in the hallway. Each niche possessed a depth of 0.8 m and a width of 0.6 m, deliberately sized to comfortably accommodate the physical footprint of the robot model (0.55$m \times$0.4$m$). To simulate realistic movement constraints during an evacuation, the participant's walking speed was capped at 2.0 m/s~\cite{wang2021} within the narrow corridor and increased to 4.0 m/s~\cite{hori2011} once they reached the spacious open hallway.

\begin{figure}[t]
     \centering
         \begin{subfigure}[b]{0.28\textwidth}
             \centering
             \includegraphics[width=\textwidth]{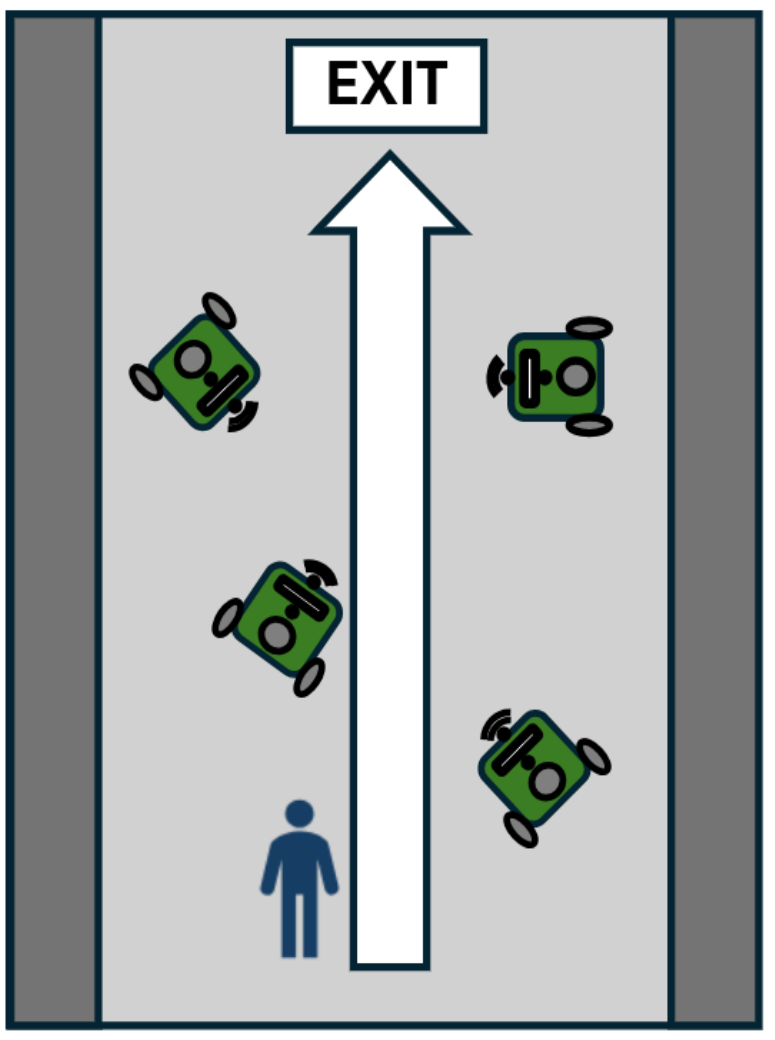}
             \caption{\textbf{Collective Freeze}}
             \vspace{2pt}
             \scriptsize{Robots halt at their initial locations.}
             \label{fig:freeze}
         \end{subfigure}
         \hspace{1pt}
         \begin{subfigure}[b]{0.28\textwidth}
             \centering
             \includegraphics[width=\textwidth]{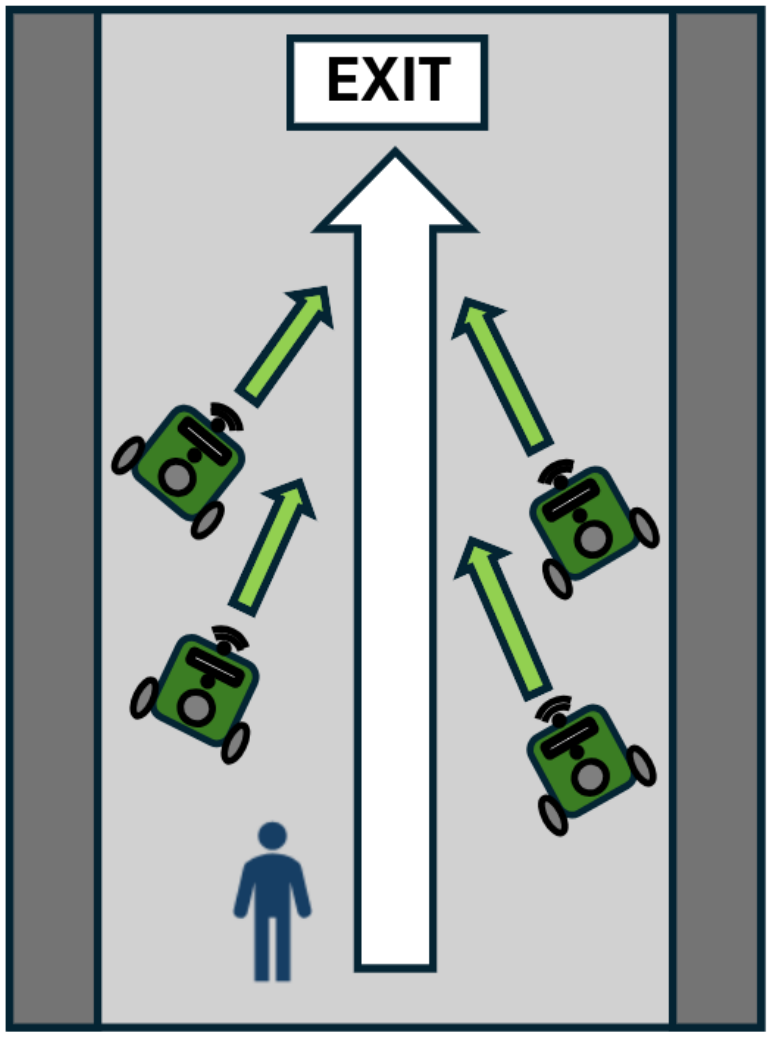}
             \caption{\textbf{Shortest-Path}}
             \vspace{2pt}
             \scriptsize{Robots navigate to the exit via the shortest route.}
             \label{fig:shortest}
         \end{subfigure}
         \hspace{1pt}
         \begin{subfigure}[b]{0.28\textwidth}
             \centering
             \includegraphics[width=\textwidth]{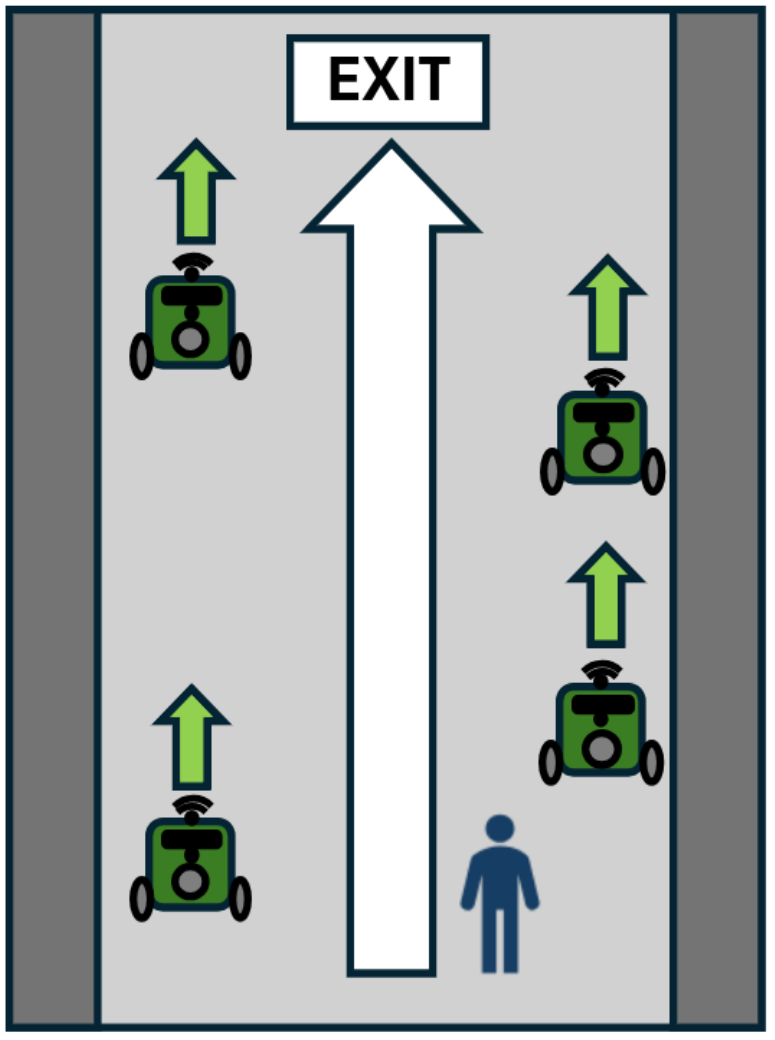}
             \caption{\textbf{LineEscape}}
             \vspace{2pt}
             \scriptsize{Robots yield, then go along the wall to the exit.}
             \label{fig:line}
         \end{subfigure}
    
         \vspace{15pt} 
    
         \begin{subfigure}[b]{0.28\textwidth}
             \centering
             \includegraphics[width=\textwidth]{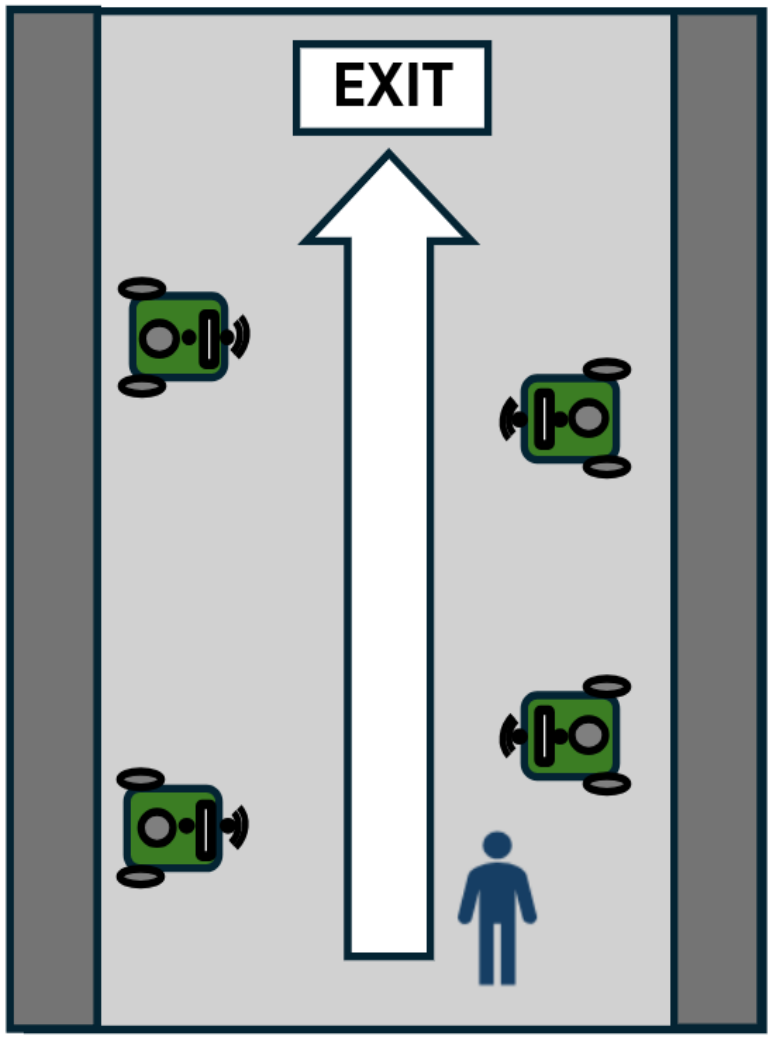}
             \caption{\textbf{Hide (No Niche)}}
             \vspace{2pt}
             \scriptsize{Robots retreat to wall as static obstacle.}
             \label{fig:hide_no}
         \end{subfigure}
         \hspace{1pt}
         \begin{subfigure}[b]{0.28\textwidth}
             \centering
             \includegraphics[width=\textwidth]{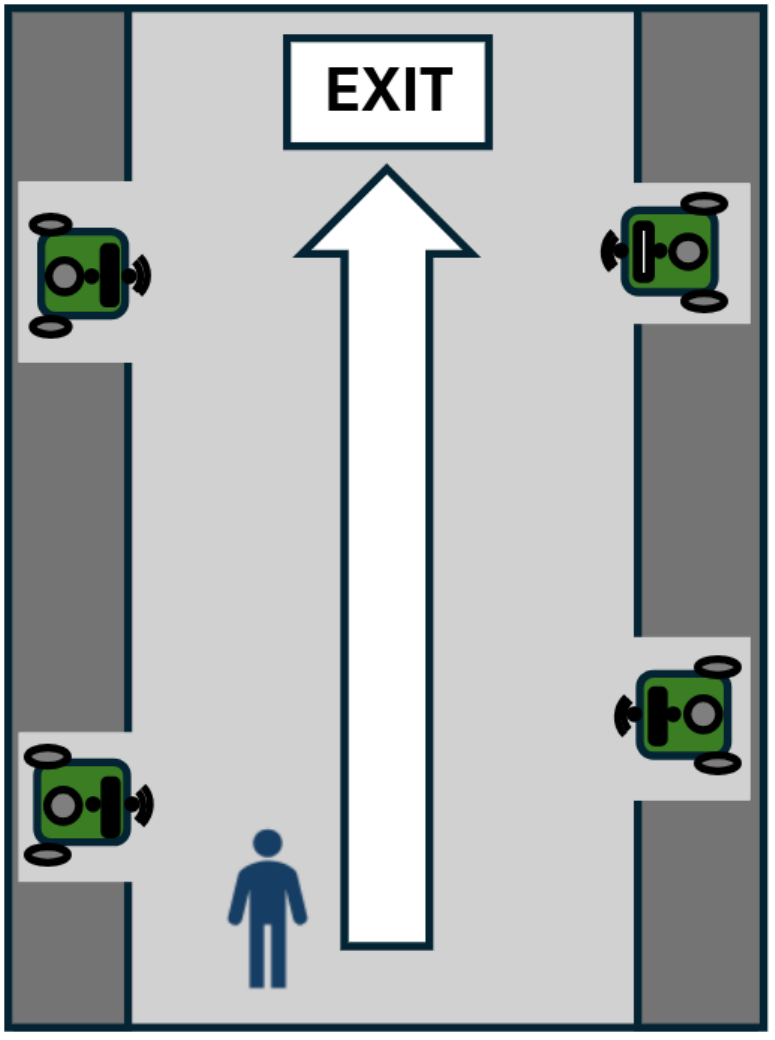}
             \caption{\textbf{Hide (with Niche)}}
             \vspace{2pt}
             \scriptsize{Robots retreat into refuge niches inside the wall.}
             \label{fig:hide_yes}
         \end{subfigure}
    \caption{Schematic of the evaluated multi-robot navigation strategies.}
    \label{fig:evacuation_strategies}
\end{figure}

The study employed a $2 \times 4$ mixed experimental design. The between-subjects factor was the \textit{Environment Condition}, manipulated by the architectural layout of the corridor: \texttt{No Affordance} (\texttt{A0}, a flat, straight corridor) versus \texttt{Affordance} (\texttt{A1}, a corridor featuring several physical refuge niches). The within-subjects factor was the \textit{Robot Strategy}. During the evacuation, participants encountered a group of four virtual autonomous mobile robots in the corridor. To isolate the effect of the navigation algorithms, the initial starting positions of these four robots were strictly identical across all strategy trials. In each trial, all four robots simultaneously executed one of four distinct social navigation behaviours (\texttt{Hide}, \texttt{ShortestPath}, \texttt{Freeze}, or \texttt{LineEscape}). The specific spatial mechanics of each strategy, including how the \texttt{Hide} behaviour adapts to the presence or absence of environmental niches, are detailed and visualised in Fig.~\ref{fig:evacuation_strategies}.

To ensure safety and prevent physical clipping in the virtual environment, the robots were equipped with an automatic obstacle avoidance safety mechanism. If a human entered within a 5.0m radius of a robot, that specific robot halved its speed; if the human breached a critical 1.5m radius, the robot stopped entirely.

\subsection{Experimental Procedure}

The experimental workflow followed a structured four-stage pipeline (Fig.~\ref{fig:pipeline}). After completing a demographic questionnaire focusing on prior robot experience, participants underwent unlimited baseline training to familiarise themselves with the controls and the hotel layout without robots present.

The formal experiment comprised four trials, one for each robot strategy, with sequences counterbalanced using a Balanced Latin Square design~\cite{williams1949} to minimize ordering effects. To induce urgency, a fire alarm triggered randomly (4–10 seconds) while the participant waited in the starting room. Upon the door automatically opening, robots initiated their navigation strategies only after the participant crossed a doorway checkpoint, effectively isolating the robot's algorithmic impact from individual human reaction times. Once the participants reached the designated safe zone in the hallway, the trial concluded, and they were prompted to evaluate the robots' overall behaviours using an in-game questionnaire. Finally, participants provided qualitative feedback and an explicit preference ranking of all four strategies after completing all trials.

\subsection{Data Collection and Participants}
Data was continuously logged during the experiment across three dimensions: subjective perception, objective trajectory metrics, and participant feedback. At the end of each trial, subjective perceptions were measured using an 8-item questionnaire on a 7-point Likert scale. These items assessed \textit{perceived obstruction and efficiency} (Q1: Obstructive-Unobstructive, Q2: Delay-No Delay), \textit{cognitive load} adapted from the NASA-TLX~\cite{hart1988} (Q3: Mental Demand, Q4: Temporal Demand), \textit{perceived safety} adapted from the Godspeed questionnaire~\cite{bartneck2009} (Q5: Anxious-Relaxed, Q6: Agitated-Calm), and \textit{social evaluation} (Q7: Selfish-Selfless, Q8: Rude-Polite). Objective human trajectory data was recorded at each frame, allowing for the extraction of spatial-temporal features. Finally, for the \textit{Moderator} analysis, participants in the \texttt{A1} group were asked a binary question regarding their \textit{Niche Awareness} (i.e., whether they consciously noticed the available refuge spaces during the evacuation).

A total of 56 participants (28 in the \texttt{A0} condition and 28 in the \texttt{A1} condition) were recruited for this study, which exceeds the minimum sample size of 44 determined by an \textit{a priori} power analysis (80\% power, $\alpha=0.05$, medium effect size Cohen's $f=0.25$) for a $2\times4$ mixed-design ANOVA. The screening process required all participants to be fluent in English and over the age of 18. Participants were recruited opportunistically in person from the libraries, robotics laboratories, and teaching areas of the University of Bristol and the University of the West of England. The participants were predominantly young adults aged between 18 and 34 (including 17 participants aged 18-24, and 39 participants aged 25-34). Regarding prior robot experience, the \texttt{A0} group was perfectly balanced (14 Plenty, 14 Little), while the \texttt{A1} group comprised 8 participants with Plenty of experience and 20 with Little experience. Participant gender was not recorded as it was deemed outside the scope of the current spatial negotiation hypothesis.  Approval for this experiment was granted by the University of Bristol Ethics Committee (ID 29326).

\section{Results}
\label{sec:results}

\subsection{Omnibus Test Results}

To evaluate the impact of robot strategy and environmental affordances on user perception, an \textit{Aligned Rank Transform Analysis of Variance (ART-ANOVA)}~\cite{jacob2011} was conducted on the eight evaluation questions and strategy-preference rankings. The analysis revealed a highly significant main effect of robot strategy across all dependent questions (all $p < 0.001$), with effect sizes ranging from medium to large (partial eta-squared $\eta_p^2$ ranging from $0.135$ to $0.571$). Furthermore, the omnibus test for participants' overall preference, measured by the strategy rank, also demonstrated a highly significant difference with a large effect size ($F(3, 162) = 92.60, p < 2 \times 10^{-16}, \eta_p^2 = 0.632$). This identifies the robot's social navigation behaviour as the primary driver of user perception. In contrast, the main effect of the environmental condition (the presence or absence of niches) was statistically non-significant across all baseline ratings with negligible effect sizes.

While environmental affordances alone did not produce a global shift in user perception, the analysis revealed significant interaction effects between the environment and robot strategy, specifically for perceived delay ($F(3, 162) = 3.94, p = 0.010, \eta_p^2 = 0.068$) and agitation ($F(3, 162) = 3.64, p = 0.014, \eta_p^2 = 0.063$). Both interactions reflect a medium effect size, indicating that the psychological and perceived impact of certain navigation strategies is meaningfully sensitive to the spatial context in which they are executed.

\subsection{Post-hoc Analysis and Contextual Effects}
\label{sec:posthoc_results}

\begin{figure}[htbp]
\centering
\includegraphics[width=\textwidth]{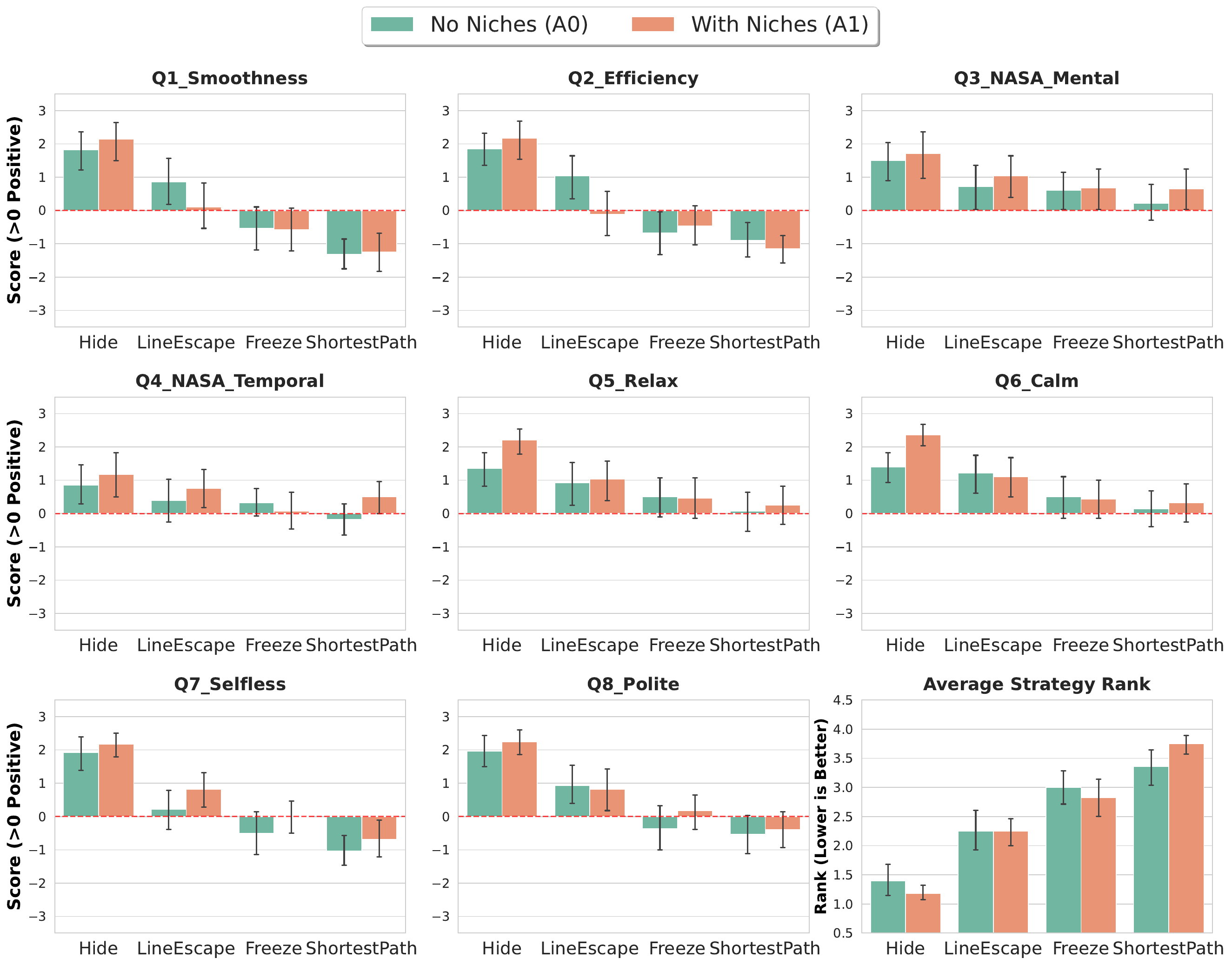}
\caption{Subjective evaluation scores (Q1-Q8) and average strategy rankings across the four robot navigation strategies, categorised by environmental condition (\texttt{No Niches} vs. \texttt{With Niches}). For Q1-Q8, higher scores indicate more positive user perceptions, while for the Strategy Rank, lower values indicate a stronger preference. Error bars represent 95\% confidence intervals calculated via bootstrapping. The \texttt{Hide} strategy demonstrates a consistent and significant advantage across multiple dimensions.}\label{fig:likert_avgrank}
\end{figure}

\begin{figure}[htbp]
\centering
\includegraphics[width=\textwidth]{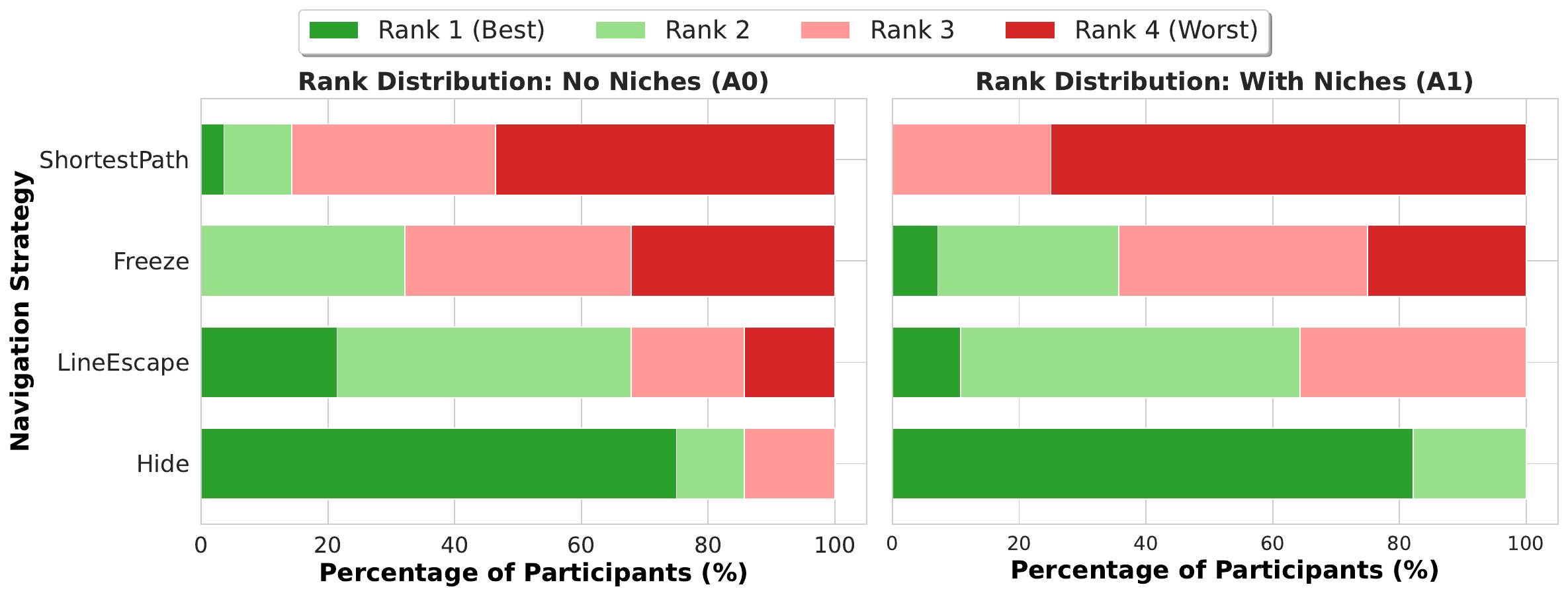}
\caption{Distribution of participants' explicit preference rankings (1 = Best, 4 = Worst) for each navigation strategy under the \texttt{No Niches} (\texttt{A0}) and \texttt{With Niches} (\texttt{A1}) conditions. The \texttt{Hide} strategy clearly dominates the first-rank choices across both environments, whereas \texttt{ShortestPath} is predominantly ranked last.} \label{fig:rank_dist}
\end{figure}

To follow up on the significant omnibus effects, we conducted pairwise contrasts using the ART methodology. To maintain a rigorous alpha level, the family-wise error rate was controlled using the Holm correction~\cite{Holm1979}. For variables without significant interaction effects (i.e., Q1, Q3, Q4, Q5, Q7, Q8, and Strategy Rank), the main effects of the robots' navigation strategy were directly analysed. As illustrated in Fig.~\ref{fig:likert_avgrank} and Fig.~\ref{fig:rank_dist}, the ranking analysis confirmed a clear, statistically significant preference hierarchy among the tested behaviours: \texttt{Hide} ($Mean=1.29$) $>$ \texttt{LineEscape}($Mean=2.25$) $>$ \texttt{Freeze}($Mean=2.91$) $>$ \texttt{ShortestPath}($Mean=3.55$) ($p < 0.001$). Overall, the \texttt{Hide} strategy significantly outperformed all other evaluated strategies in mitigating negative user experiences across these dimensions ($p < 0.001$, with effect sizes $r$ ranging from $0.28$ to $0.78$). Conversely, the \texttt{ShortestPath} approach was ranked the lowest and was consistently perceived as the most obstructive, selfish, and anxiety-inducing. Due to the extensive number of pairwise comparisons, detailed statistical values are available in our supplementary material GitHub repository.

\begin{figure}[htbp]
\centering
\includegraphics[width=\textwidth]{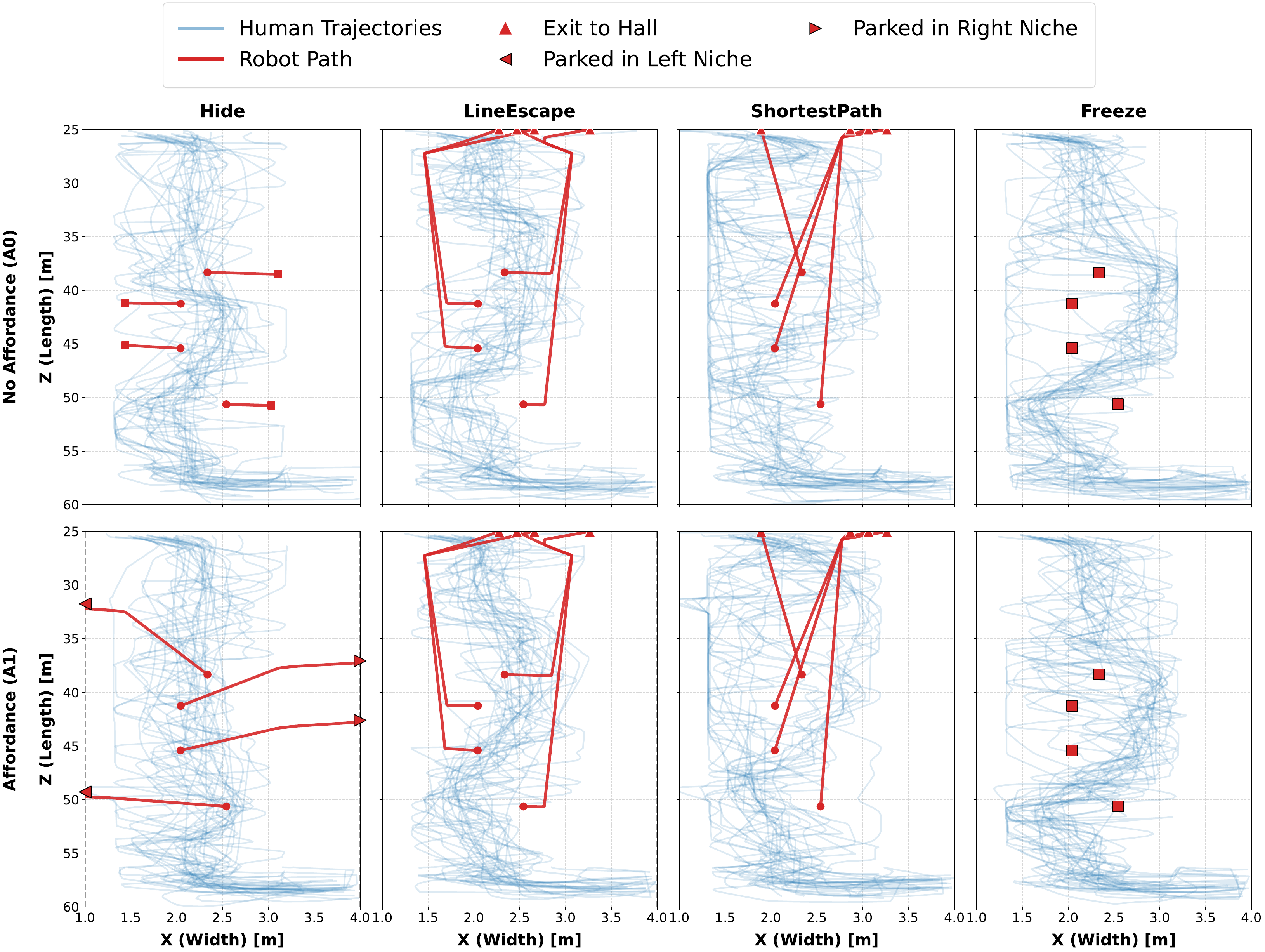}
\caption{Movement paths of all 28 human participants (blue lines) and the robots (red lines) for each of the different navigation strategies and environmental conditions.} \label{fig:trajectory_vis}
\end{figure}

The variables Q2 (Efficiency) and Q6 (Calm) were excluded from the pooled main effect analysis because their significant Condition $\times$ Strategy interactions make generalised main effects misleading. Instead, a simple main effects analysis was performed to unpack how environmental context modulated these specific perceptions. We found that when the \texttt{Hide} strategy was employed in environments with niches, participants reported being significantly calmer ($t(145) = 2.29, p = 0.023, r = 0.19$) and experienced a marginal trend increase in perceived efficiency ($t(177) = 1.93, p = 0.055, r = 0.14$) compared to the no-niche condition. 

\begin{figure}[htbp]
\centering
\includegraphics[width=\textwidth]{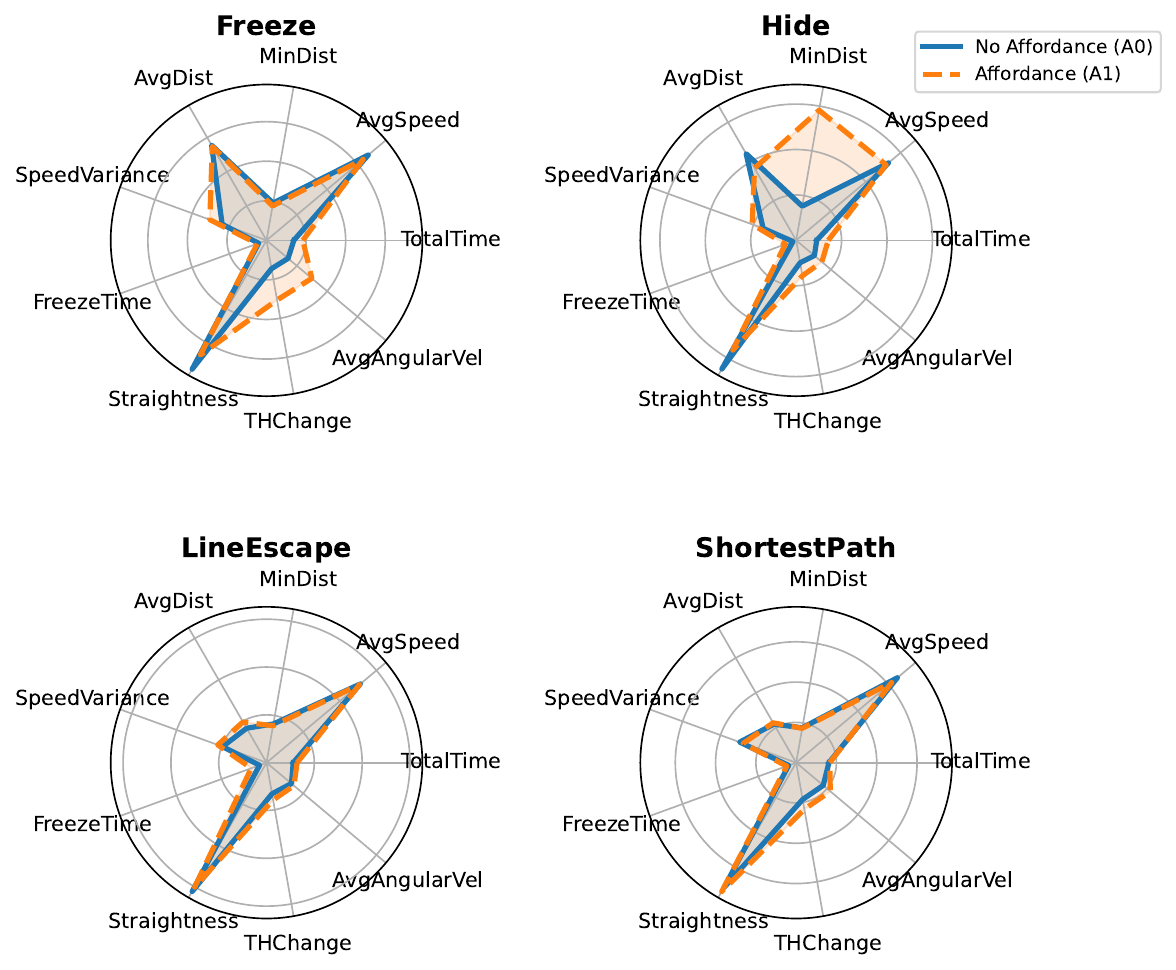}
\caption{Radar charts comparing nine objective spatial-temporal trajectory metrics across the four navigation strategies between the \texttt{No Affordance} (\texttt{A0}) and \texttt{Affordance} (\texttt{A1}) conditions. Notably, the \texttt{Hide} strategy exhibits a substantial increase in the minimum clearance distance (MinDist) when niches are present, physically validating its effectiveness in mitigating perceived obstruction.} \label{fig:metrics_radar}
\end{figure}

This psychological relief is supported by objective human trajectory data, visually presented in Fig.~\ref{fig:trajectory_vis}. To quantify the physical interaction dynamics, we extracted a comprehensive set of spatial-temporal features from the human trajectories. These metrics included minimum clearance distance (\texttt{MinDist}), average distance to the robot (\texttt{AvgDist}), path straightness (\texttt{Straightness}), average speed (\texttt{AvgSpeed}), speed variance (\texttt{SpeedVariance}), total time (\texttt{TotalTime}), freeze time (\texttt{FreezeTime}), total heading change (\texttt{THChange}), and average angular velocity (\texttt{AvgAngularVel}). Because human trajectory data typically violates normality assumptions, we employed robust non-parametric statistical methods on 54 valid data points (\texttt{A0}: 26, \texttt{A1}: 28), excluding two participants due to data logging errors.
Between-subjects comparisons were conducted using the Mann-Whitney U test. Within-subjects comparisons utilised the Friedman test for overall variance, followed by post-hoc Wilcoxon signed-rank tests. To rigorously control the family-wise error rate during pairwise comparisons, all post-hoc $p$-values were adjusted using the Holm correction.

As depicted in the radar charts (Fig.~\ref{fig:metrics_radar}), between-subjects analysis revealed that the \texttt{Hide} strategy afforded a substantially larger minimum clearance distance (MinDist) when niches were present compared to the no-niche condition (1.66m vs. 0.81m, $p < 0.001$, Effect Size = 0.85). Although participants exhibited slightly lower path straightness (0.87 vs. 0.92, $p = 0.049$) and a lower average distance to the robot (3.62m vs. 4.08m, $p < 0.001$) in the affordance condition, these kinematic deviations did not diminish their overarching sense of safety.

In contrast, applying the \texttt{LineEscape} strategy in the presence of niches had an adverse effect, significantly decreasing perceived efficiency ($t(177) = -2.08, p = 0.038, r = 0.15$). Crucially, our trajectory analysis showed no significant between-subjects differences in objective trajectory metrics for \texttt{LineEscape} across the two environments, suggesting that this subjective penalty originates from unfulfilled cognitive expectations regarding niche utilisation, rather than actual physical impedance. A further asymmetry in the data shows the \texttt{Freeze} and \texttt{ShortestPath} strategies do not suffer a significant drop in perceived efficiency when they fail to use an available niche.

\subsection{Moderator Analysis and Robustness Check}

To further understand how internal participant factors influenced user perceptions and to validate the robustness of our overall findings, we examined three potential moderators: niche awareness, age group, and prior robot experience (Plenty vs. Little). We employed \textit{Cumulative Link Mixed Models (CLMM)}~\cite{rune2025} treating the Likert data as ordinal, while strictly controlling for age and overall robot experience as covariates. Neither age group nor overall prior experience exhibited significant global main effects across the metrics ($p > 0.05$), and the CLMM results fully preserved the significant main effects of strategy and localized interaction effects. The non-significant global effect of robot experience in the CLMM confirms that the between-subjects imbalance in experience does not affect our primary findings, and that our results are robust across statistical approaches rather than artifacts of specific demographic biases.

While prior experience did not globally skew the results, it did exhibit specific localised moderating effects. To evaluate these, we conducted post-hoc Wilcoxon rank-sum tests, applying the Holm~\cite{Holm1979} correction within each evaluation metric across the four navigation strategies to control for Type I errors. The analysis revealed that participants with prior robot experience perceived the \texttt{Hide} strategy as significantly less obstructive ($p_{holm} = 0.032$, raw $p = 0.008$, Mean = 1.32 vs. 2.41) and demonstrated a trend toward reduced anxiety during its execution (raw $p = 0.069$, Mean = 1.32 vs. 2.09) compared to novices.

\section{Discussion and Future Work}
\label{sec:discussion}

\subsection{The Hierarchy of Proactive Space-Yielding in Confined Spaces}
Our results establish a clear and statistically robust preference hierarchy for single-person interactions in narrow environments: \texttt{Hide} $>$ \texttt{LineEscape} $>$ \texttt{Freeze} $>$ \texttt{ShortestPath}. As comprehensively supported by the subjective evaluation metrics across all eight dimensions, the \texttt{Hide} strategy consistently outperformed all alternatives. It proved to be the most effective approach for mitigating perceived obstruction, cognitive load, and agitation. The preference for this proactive space-yielding behaviour underscores the necessity of communicating clear social intent in high-pressure or confined scenarios~\cite{dragan2013}. In contrast, the \texttt{ShortestPath} strategy, which prioritises kinematic efficiency over social spatial negotiation, was universally perceived as the most obstructive and selfish. This confirms that in confined spaces, pure flow efficiency must be subordinated to human-centric spatial comfort.

\subsection{Environmental Affordance and Expectation Violation}
A key contribution of this research is revealing the dynamic interplay between robot behaviour and spatial semantics. We found that environmental affordances (i.e., refuge niches) shape human psychological expectations. As demonstrated by the trajectory features in Fig.~\ref{fig:metrics_radar}, when a robot proactively utilises a niche (\texttt{Hide}), it moves out of the human's personal space into social space~\cite{hall1966}. This physical yielding validates human spatial expectations, resulting in amplified psychological calmness.

Conversely, our findings regarding the \texttt{LineEscape} strategy suggest a phenomenon consistent with Expectation Violation~\cite{burgoon1976}. While \texttt{LineEscape} and \texttt{ShortestPath} may be beneficial for macroscopic crowd flow, they fail in single-person interactions when clear physical refuges exist. Specifically, as detailed in Sec.~\ref{sec:posthoc_results}, participants significantly penalised the \texttt{LineEscape} strategy for perceived delay in the \texttt{Affordance} condition ($t(177) = -2.08$, $p = 0.038$, $r = 0.15$), even though objective trajectory metrics in Fig.~\ref{fig:metrics_radar} confirmed that no actual physical deviation or obstruction occurred.

We propose that this apparent expectation violation is modulated by the robot's implied social competence. While the \texttt{Freeze} and \texttt{ShortestPath} strategies also ignore available niches, they do not incur this psychological penalty. This may be because both strategies have already signaled a lack of spatial awareness or social agency—\texttt{Freeze} by simply halting, and \texttt{ShortestPath} by selfishly prioritising its own kinematic efficiency —effectively establishing a severely low expectation baseline~\cite{burgoon1976}. Missing an affordance merely confirms these low expectations, acting as a floor effect where the behaviour cannot be penalised further. \texttt{LineEscape}, however, raises the observer's expectation threshold through its purposeful, active movement. Ultimately, this reveals a design trade-off: algorithms that communicate competence through active navigation inadvertently raise the social expectations they must satisfy, making the psychological cost of ignoring an affordance proportional to the robot's implied spatial agency.

\subsection{Moderating Factors: User Experience and Cognitive Awareness}

Our moderator analysis introduces a promising perspective on the long-term deployment of social robots in public spaces. As established in the results, prior robot experience significantly improved users' reception of the \texttt{Hide} strategy. It is important to note that novices still perceived the \texttt{Hide} strategy positively, rating it as unobstructive (mean = 2.41, below the neutral threshold of 4.0). However, experienced users appreciated this proactive spatial yielding significantly more (mean = 1.32), indicating they are more adept at quickly decoding complex social intents. While novices might experience a mild novelty effect~\cite{leite2013} when encountering a robot suddenly deviating into a niche, this did not compromise their overall experience. Despite the \texttt{A1} condition being disproportionately comprised of novices, the \texttt{Hide} strategy remained the overwhelming preference. This demonstrates the robust design of proactive space-yielding: it remains highly effective and intuitive for first-time users, while its perceived social benefits further amplify as the public becomes more accustomed to human-robot interactions.

Furthermore, the moderator analysis of niche awareness provided unexpected but revealing insights into human spatial cognition. The absence of a significant moderating effect suggests that a participant's conscious awareness of the niches did not dictate their psychological comfort. We posit that this aligns with Gibson's original conceptualisation of affordances~\cite{gibson1977,wolfe2004}: such spatial expectations operate at a pre-attentive cognitive level. In emergency scenarios, humans process and react to the available spatial volume intuitively, experiencing psychological friction when a robot violates these implicit spatial rules, even without requiring conscious deliberation about the environment's architectural details.

\subsection{Limitations and Future Work}

While this study provides robust insights, we acknowledge several methodological and ecological limitations. 

First, Expectation Violation was not explicitly quantified using psychometric scales. Direct measurement of expectation was deliberately excluded to prevent survey fatigue, as participants were already tasked with evaluating multiple dimensions immediately following short, high-stress evacuation trials. Instead, our interpretation relies on a classical \textit{behavioural dissociation} argument. Specifically, the psychological penalty (increased perceived delay) was localised exclusively to the \texttt{LineEscape} strategy in the affordance (\texttt{A1}) condition, despite objective trajectory metrics confirming no actual physical deviation or obstruction occurred. Furthermore, the non-significant main effect of the environment effectively rules out perceptual complexity as a confounding variable. Together, this robust divergence between subjective psychological penalty and objective physical reality provides compelling indirect evidence \textit{consistent with} a negative expectancy violation.

Second, this study purposefully scoped its investigation to micro-level, single-person interactions with the robot swarm. Navigation dynamics (such as the actual utility of the \texttt{LineEscape} strategy) may shift in high-density multi-agent evacuation scenarios, where crowd pressure and collective behaviour introduce new macro-level variables. Establishing this micro-level behavioural baseline was a necessary first step before introducing crowd confounds into future ecological validations.

Third, regarding demographics, our participant pool predominantly consisted of young adults (i.e., 18--34) from a university setting. Given that mobile service robots are frequently deployed in public infrastructure, future studies must broaden this scope to include older adults and children, whose distinct mobility constraints and baseline spatial expectations could alter how they react to yielding behaviours under stress.

Finally, it remains unclear to what extent self-reported sentiments in a virtual experiment fully resemble real-world emergencies. Factors inherent to physical embodiment, such as mechanical motor noise and genuine collision risks, could significantly shift a user's baseline expectations and amplify the psychological friction experienced, necessitating future real-world validations with physical robot swarms.

Future work will focus on addressing these gaps. From an evaluation standpoint, future studies should incorporate diverse demographics and dedicated Expectation Violation scales (e.g., the RoSAS scale~\cite{carpinella2017}) alongside physical robot trials. Algorithmically, we aim to integrate these insights into a context-aware navigation stack that dynamically maps environmental affordances in real-time, allowing robots to autonomously switch between macro-level flow optimisation and micro-level proactive yielding based on crowd density and spatial semantics.

\subsection{Broader Social Implications and Design Principles}

Our findings suggest that as robots integrate into public infrastructure, their "passive safety" must transition from a purely physical definition to a sociotechnical one. In emergency contexts, a robot's failure to respect spatial semantics (such as ignoring available refuge niches) does not merely create a physical obstacle; it generates psychological friction and cognitive load for evacuees. This implies that for socially-aware navigation, robots should be programmed with a hierarchy of spatial obligations where \textit{yielding space} is prioritised over \textit{trajectory efficiency}. For designers, this means implementing \textit{semantic-aware} navigation stacks~\cite{kostavelis2015,alqobali2023} that can dynamically identify environmental affordances in real-time, ensuring that robot swarms proactively remove themselves from the human personal space during high-stress scenarios to maintain both physical flow and psychological calm~\cite{hall1966}.

\section{Conclusion}
\label{sec:conclusion}

This study investigated human psychological perceptions of multi-robot yielding strategies during confined corridor evacuations, moving beyond traditional collision avoidance to incorporate the crucial dimension of environmental affordances. Through a highly controlled game-based experiment, we evaluated four distinct robot swarm behaviours across environments with and without physical refuge niches. 

Our findings established a clear preference hierarchy (\texttt{Hide} $>$ \texttt{LineEscape} $>$ \texttt{Freeze} $>$ \texttt{ShortestPath}), demonstrating that proactive space-yielding is paramount for mitigating perceived obstruction and cognitive load. Most importantly, we showed that the presence of spatial affordances strongly shapes human cognitive expectations. On the positive side, when a robot proactively utilises physical niches (\texttt{Hide}), it significantly amplifies users' psychological relief and objective physical clearance. Conversely, when flow-compliant algorithms like \texttt{LineEscape} fail to utilise these obvious niches in micro-level interactions, it likely triggers Expectation Violations, consistent with the observed reduction in reported human comfort. Additionally, we found that prior robot interaction experience helps mitigate these negative perceptions, highlighting a learning curve in human decoding of complex social intents.

This research underscores a principle for Human-Robot Interaction in high-stakes contexts: passive safety cannot be achieved through physics alone. Future multi-robot social navigation algorithms must evolve from merely being \textit{obstacle-aware} to becoming \textit{semantic-aware}. Only by seamlessly aligning robot behaviours with local spatial semantics and human psychological expectations can we ensure safe, efficient, and socially acceptable human-robot coexistence.

\begin{credits}
\subsubsection{\ackname} This work was supported by the China Scholarship Council (CSC No. 202408060197). Generative AI tools were used for language editing and code generation; the authors verified all content and assume full responsibility for this publication.

\end{credits}
%
%
%
\bibliographystyle{splncs04}
\bibliography{finalicsr26}
%





\end{document}